\documentclass[conference]{IEEEtran}
\IEEEoverridecommandlockouts
\usepackage{cite}
\usepackage{amsmath,amssymb,amsfonts}
\usepackage{algorithmic}
\usepackage{graphicx}
\usepackage{textcomp}
\usepackage{multirow}
\usepackage[table]{xcolor}
\usepackage{balance}
\usepackage{color}

\usepackage[algo2e,ruled]{algorithm2e}
\usepackage{amsthm}
\newtheorem{theorem}{\hspace{0pt}\bf Theorem}

\DeclareMathOperator*{\argmin}{arg\,min}

\def\BibTeX{{\rm B\kern-.05em{\sc i\kern-.025em b}\kern-.08em
    T\kern-.1667em\lower.7ex\hbox{E}\kern-.125emX}}
\begin{document}

\title{Sufficiently Accurate Model Learning}

\author{Clark Zhang$^{1}$, Arbaaz Khan$^{1}$, Santiago Paternain$^{2}$, Alejandro Ribeiro$^{2}$ 
\thanks{$^{1}$Authors are with the GRASP Lab and the    $^{2}$Electrical and Systems Engineering Department, University of Pennsylvania, USA. \{clarkz, arbaazk, spater, aribeiro\}@seas.upenn.edu}
}

\maketitle

\begin{abstract}
Modeling how a robot interacts with the environment around it is an important prerequisite for designing control and planning algorithms. In fact, the performance of controllers and planners is highly dependent on the quality of the model. One popular approach is to learn data driven models in order to compensate for inaccurate physical measurements and to adapt to systems that evolve over time. In this paper, we investigate a method to regularize model learning techniques to provide better error characteristics for traditional control and planning algorithms. This work proposes learning ``Sufficiently Accurate" models of dynamics using a primal-dual method that can explicitly enforce constraints on the error in pre-defined parts of the state-space. The result of this method is that the error characteristics of the learned model is more predictable and can be better utilized by planning and control algorithms. The characteristics of Sufficiently Accurate models are analyzed through experiments on a simulated ball paddle system.
\end{abstract}
\begin{IEEEkeywords}
Model Learning, Planning, Control
\end{IEEEkeywords}

\section{Introduction}
One of the fundamental problems in robotics is the design of controllers and planners for complex dynamical systems. These algorithms rely on models of robots that are derived from physical laws using measured physical constants. These measurements may not be accurate, since the robot and its environment may change over time, resulting in a degradation of performance of the control and planning algorithms. Recent works address these errors in estimation by using data driven models to adapt an initial analytic model \cite{lee2017gp, williams2016information}.

One popular method of using data in control and planning systems is to learn the control inputs directly whether through reinforcement-learning algorithms \cite{levine2013guided} \cite{nair2018visual}, imitation learning \cite{ross2011reduction} \cite{khan2017memory}, or other means. This may work for specific tasks, but can have difficulty adapting to different tasks or task parameters such as different control constraints. Learning a model of the system is therefore more flexible and can be used with a variety of existing algorithms.

Many modern controllers and planners rely on solving optimization problems such as iLQR \cite{tassa2014control}, CHOMP \cite{ratliff2009chomp}, and TrajOpt \cite{schulman2013finding}. These methods require differentiable forward dynamics models that have well behaved gradients in addition to being accurate. This paper formulates the model learning problem as a constrained optimization problem that seeks to provide such algorithms with predictable errors that enable them to perform well in a variety of scenarios.

\begin{figure}[t]
	\centering
	\includegraphics[width=0.35\textwidth]{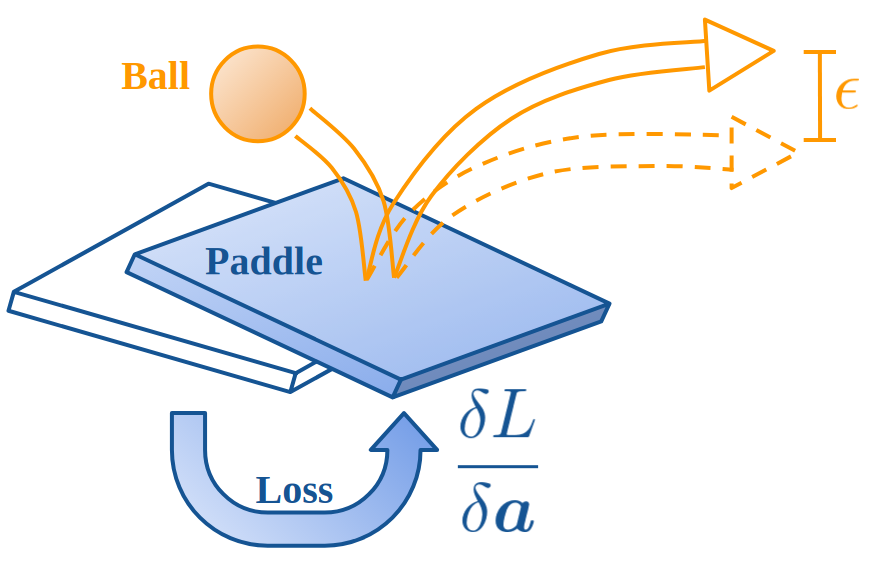}
	\caption{\textbf{Optimizing a sufficiently accurate model:} The solid line represents the true trajectory of the ball, while the dotted lines represent predicted trajectories. The paddle actions, $a$, are optimized with gradient descent on a defined loss function. For the task of bouncing the ball consistently, a model that guarantees prediction error within some bound $\epsilon$ may be sufficient. 
	}
	\label{fig:main}
\end{figure}

Learning a model is fundamentally different from learning a controller in that the controller is an end in itself but a model is useful as an intermediate step to learn a controller. Therefore, learning a model with arbitrary accuracy is not necessary. Rather, we want to learn a model that is sufficiently accurate for controller design. This paper formulates the problem of learning a model as a constrained optimization problem in which the required accuracy of the model is imposed as a set of constraints (Section \ref{sec:model_learning}). The constraints that are imposed in the model accuracy are intended to ensure that the model is {\it sufficiently accurate} for a variety of control tasks. An additional advantage derived from our problem formulation is that the accuracy constraints that are imposed in model learning allow for trading off the accuracy of the model in different parts of the state space. For example, in Fig. \ref{fig:main} we consider the problem of determining the dynamical trajectories followed by a ball that is hit by a paddle in order to design a controller that would allow us to keep the ball in the air by repeatedly hitting it when the vertical position crosses a certain threshold. We argue that for this problem it is advantageous to learn models that predict with an accuracy dependent on the velocity of the ball.

\subsection{Contributions}
This paper proposes a constraint-based formulation for learning and controlling dynamical systems. The contributions of this paper are: 
(i) a novel constrained objective function for model learning with neural networks and the adjoining constrained optimization problem for learning the controller, and (ii) a primal-dual method to solve both these problems that has small duality gap. The method is evaluated on a simulated ball bouncing task with varying task parameters and injected errors.

\section{Related Work}
\label{sec:relatedwork}
The idea of learning a model from data and using it to control systems is not new. PILCO \cite{deisenroth2011pilco} learns a probabilistic forward model with Gaussian Process Regression and is later extended to Bayesian Neural Networks \cite{gal2016improving}. Guided Policy Search \cite{levine2014learning,levine2016end} uses a Gaussian Mixture Model as a probabilistic dynamics model. Both formulate an optimization problem with the task of maximizing an expected reward. This allows a policy to be trained by backpropagating through the forward model. \cite{jordan1992forward} and \cite{gillespie2018learning} both learn neural network models. The latter then formulates a convex optimization problem by linearizing the neural network. However, it is has been observed that linearizing highly nonlinear systems often performs poorly \cite{diehl2005real}.
The aforementioned methods learn models with an objective function similar to 
\begin{equation}
\label{eq:model_learning}
\min_\theta \mathbb{E}\left\| \phi_\theta(s, a) - f(s, a) \right\|
\end{equation}
where $f(s, a)$ represents the true model dynamics and $\phi_\theta(s, a)$ is the learned model.
\cite{byravan2017se3} designs a specific neural network architecture for their forward model which uses a normalized objective. However, there are slack terms introduced to avoid numerical instability caused by dividing with small numbers. \cite{amos2018differentiable} presents a way to differentiate through the controller so that a model can be learned end to end. This method requires that the policy has converged to a fixed point which can be hard to achieve in complicated systems.

Learning models is also of great interest in reinforcement learning where a forward model can increase sample efficiency~\cite{sutton1990integrated,levine2014learning}. In addition, it has been found that learning forward and inverse models can provide additional rewards to help train a reinforcement learning agent~\cite{pathak2017curiosity}.
\cite{achiam2017constrained} introduced a policy search method for reinforcement learning that can impose expectation constraints on states and actions. 

There has also been some work in multi task learning to obtain task agnostic policies. One way to do this is with meta learning algorithms such as MAML \cite{finn2017model} or \cite{cervino2019meta} which learn policies that can be adapted to different task parameters. \cite{pathak2018zero,nair2018visual} learn policies that include a goal as an input to encourage the policy to generalize across different goals. These methods work for small numbers of task parameters but have difficulty scaling up as each additional parameter added to a policy will decrease the sample efficiency of the algorithm.

\section{Constrained Model Learning}
\label{sec:model_learning}
In this work, we consider a discrete forward dynamic model. Formally, let $\mathcal{S}$ and $\mathcal{A}$ denote the state and action spaces respectively. Then, the dynamic model is defined by a function $f:\mathcal{S}\times\mathcal{A}\to \mathcal{S}$ whose inputs are the state and action at time $n$, denoted by $s_n\in\mathcal{S} $ and $a_n\in\mathcal{A}$, and whose output is the state at time $n+1$
\begin{equation}
{s}_{n+1} = f({s}_n, {a}_n).
\end{equation}
We denote $\phi_\theta : \mathcal{S} \times \mathcal{A} \to \mathcal{S}$ as the neural network approximation of the true model, where $\theta\in\mathbb{R}^N$ represents the network parameters.
The classical approach to model learning consists of finding the parameters $\theta$ that minimize the expectation of a loss function $l:\mathcal{S}\times\mathcal{A}\times\mathbb{R}^N\to\mathbb{R}$, this is
\begin{equation}
\min_\theta  \mathbb{E}_{( {s},  {a}) \sim D} l(s,a,\phi_\theta),
\end{equation}
where $D$ denotes the sampling distribution over the state-action space. Note that $D$ is not influenced by $\theta$ and is simply a training distribution. This simple objective does not allow any control over how errors are distributed. To address this limitation, we formulate the problem of learning a model as a constrained optimization problem in which the required accuracy of the model is imposed as a set of constraints. The constraints that are imposed in the model accuracy are intended to ensure that the model is sufficiently accurate, making it suitable for a variety of control tasks. To that end, we consider $m\in \mathbb{N}$ subsets, $K_i \subseteq \mathcal{S} \times \mathcal{A}$ with $i=1,\ldots,m$, of the state-action space and constraints $h_i:\mathcal{S}\times\mathcal{A}\times\mathbb{R}^N\to \mathbb{R}$, where each component of the constraint is imposed on a different region of the space-action space. With these definitions, we propose the following optimization problem
\begin{align}\label{eq:constrained_model_learning}
P^*_\theta &=\min_\theta  \mathbb{E}_{( s,  a) \sim D} l(s,a,\phi_\theta) \mathbb{I}_{\mathcal{K}_0}(s,a) \\
& \text{s.t. } \mathbb{E}_{(s,a) \sim D} h_i(s,a,\theta) \mathbb{I}_{\mathcal{K}_i}(s,a) \leq \epsilon_i \quad \forall i\ldots m, \nonumber
\end{align}
where $\mathbb{I}_{\mathcal{K}_i}(s,a)$ with $i=0,\ldots, m$ are indicator functions taking the value $1$ if $(s,a)\in\mathcal{K}_i$ and $0$ otherwise. For simplicity we define $g_i(s,a,\phi_{\theta}) = h_i(s,a,\phi_{\theta})\mathbb{I}_{\mathcal{K}_i}-\epsilon_i$ for all $i=1,\ldots ,m$. In the next section we present a primal dual algorithm to solve the optimization problem \eqref{eq:constrained_model_learning}. Before doing so, we consider a specific case to illustrate the sufficiently accurate learning framework. 

\textbf{Example: Normalized Error Model.}
As a specific case of the previous formulation, we consider the minimization of a normalized error. The error tolerance in a forward model is directly related to the magnitude of the quantity being estimated. For example, 1 unit of error when the output is 100 is different from 1 unit of error when the output is 1. A natural way to mitigate these difference is to consider a normalized error, $\left\| \phi_\theta( {s},  {a}) - f( {s},  {a}) \right\| / \left\| f( {s},  {a}) \right\|$. It is often the case that small values of $f( {s},  {a})$ are hard to even measure accurately, thus, it matters only that the error is bounded rather than minimized. The set of non-small values will be defined as $\mathcal{K} = \{ ( {s},  {a}) : \left\| f( {s},  {a}) \right\| \geq \delta \}$, where $\delta > 0$. One can then pose the following model learning objective
\begin{align}\label{eq:constrained_model_learning_example}
\min_\theta & \mathbb{E}_{( {s},  {a}) \sim D} \frac{\left\| \phi_\theta( {s},  {a}) - f( {s},  {a}) \right\|}{ \left\| f( {s},  {a}) \right\| } \mathbb{I}_{\mathcal{K}} \\
& \text{s.t. } \mathbb{E}_{( {s},  {a}) \sim D} \left\| \phi_\theta( {s},  {a}) - f( {s},  {a}) \right\| \mathbb{I}_{\mathcal{K}^C}  \leq \epsilon . \nonumber
\end{align}
This objective simply states that for all large enough $f( {s},  {a})$, the normalized error should be minimized, and all small values should be bounded by $\epsilon$. This constraint can be seen as a regularization placed on the model learning objective to avoid overfitting the model to small valued labels. 

\begin{figure}[h]
	\centering
	\includegraphics[scale=0.6]{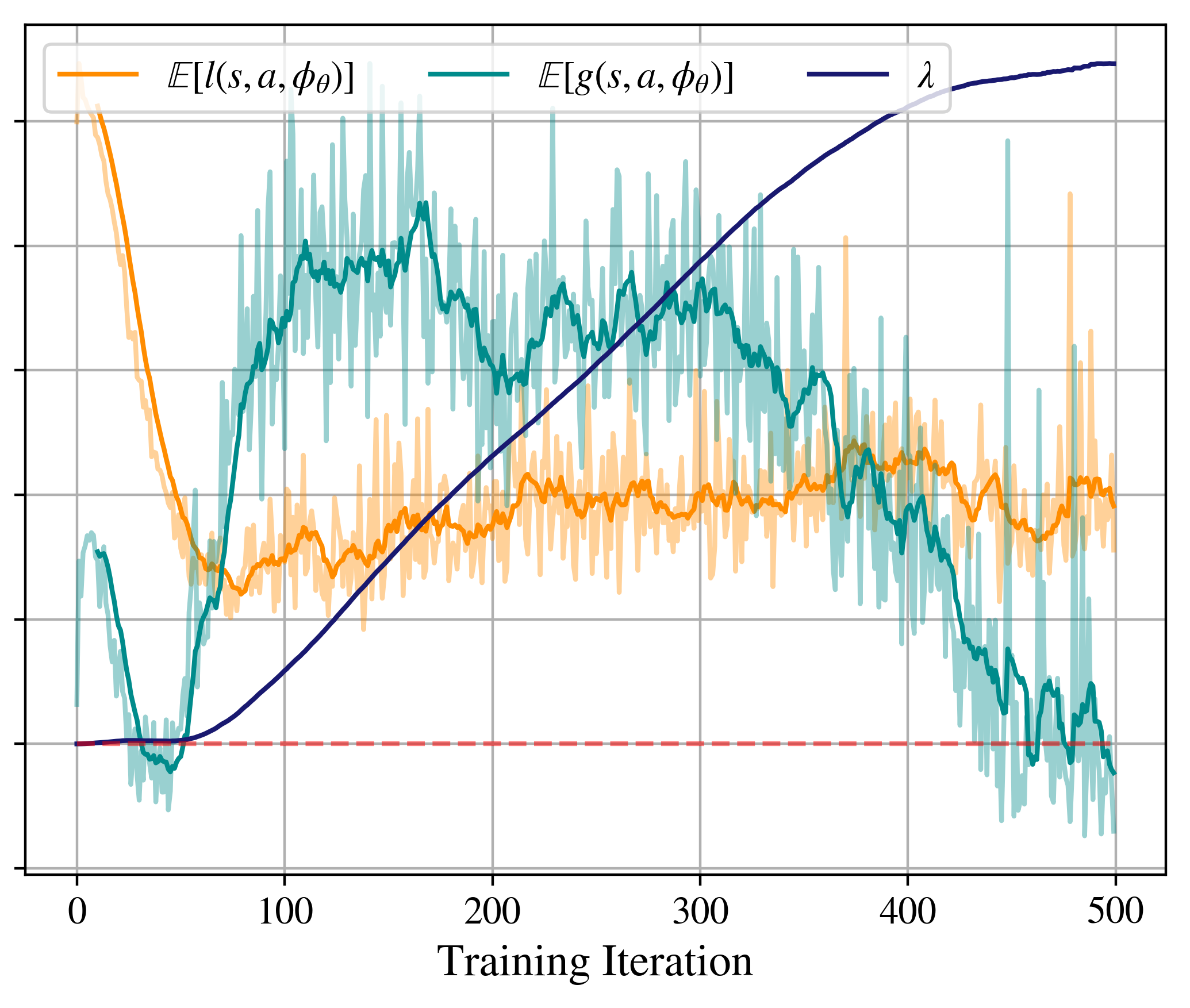}
	\caption{\textbf{Model training.} The orange curve shows the value of the objective function, $l$. Teal curve shows the value of the constraint function, $g$. Dark blue curve shows the dual variable, $\lambda$. The objective and constraint functions are smoothed and shown in a darker color. The dotted red line shows the $y=0$ line and the constraint curve must go below that for the problem to have a feasible solution. The curves have different values and are normalized so that they can be displayed on one graph.}
	\label{fig:model_learning}
\end{figure}

\subsection{Primal-Dual algorithm}\label{Methodology}
The problem of sufficiently accurate learning can be formulated as the constrained optimization problem \eqref{eq:constrained_model_learning}. A possible approach to solve said problem is through primal-dual methods. Let us start by defining a vector of multipliers $\lambda\in\mathbb{R}^P_+$ and the Lagrangian associated to problem \eqref{eq:constrained_model_learning}  
\begin{equation}
L(\theta, \lambda) = \mathbb{E}[ \mathcal{L}( {s},  {a}, \phi_\theta)] + \lambda^\top \mathbb{E} [g( {s},  {a}, \phi_\theta)].
\end{equation}
where to simplify notation we have defined $\mathcal{L}( {s},  {a}, \phi_\theta)  = l(s,a,\phi_\theta)\mathbb{I}_{\mathcal{K}_0}$ and we have dropped the distribution $D$. The Lagrangian allows us to define the dual problem as
\begin{align}\label{eq:opt_problem_dual}
D_\theta^* &= \max_\lambda \min_\theta L(\theta, \lambda)\\
&  \text{s.t. } \lambda \succeq 0 \nonumber. 
\end{align}
The duality gap is defined by the difference $P_\theta^* - D_\theta^*$. When an optimization problem has zero duality gap (we show in Section \ref{sec_zero_duality_gap} that the problem of learning sufficiently accurate models has close to zero duality gap), then the solutions of both optimization problems in  \eqref{eq:constrained_model_learning} and \eqref{eq:opt_problem_dual} are the same. The optimal primal variable, $\theta^*$, must necessarily minimize $L$ given the optimal dual variable, $\lambda^*$. Likewise, the optimal dual variable must maximize the Lagrangian given the optimal primal variable. This leads to the widely used primal-dual method, where the Lagrangian is iteratively minimized with respect to the primal variable and maximized with respect to the dual. This minimization/maximization can be solved by computing gradient descent steps with respect to $\theta$ and gradient ascent steps with respect to $\lambda$. The gradient of the Lagrangian with respect to $\theta$ takes the form
\begin{align}
\nabla_\theta L(\theta, \lambda) &= \mathbb{E} [\nabla_\theta \mathcal{L}( {s}, {a}, \phi_\theta)] +\sum_{i=1}^m \lambda_i \mathbb{E} [\nabla_\theta g_i( {s}, {a}, \phi_\theta)], 
\end{align}
and the gradient of the Lagrangian with respect to the multiplier $\lambda$ yields
\begin{equation}
\nabla_\lambda L(\theta, \lambda) = \mathbb{E} [g( {s}, {a}, \phi_\theta)]. 
\end{equation}
Each iteration must be followed by a projection of $\lambda$ onto the positive orthant to make sure it is non negative. Given a static distribution of states and actions to optimize this model, the algorithm is summarized in Algorithm \ref{alg:model_learning}.
\begin{algorithm2e}[h]
	\label{alg:model_learning}
	\SetAlgoLined
	\KwIn{$\theta_0$, initial model parameters\\
		\quad \quad \quad $D$, dataset of $( {s},  {a}, f( {s},  {a}))$ tuples}
	$\lambda \leftarrow \lambda_0$ , set dual variable to an initial value  \;
	$\theta \leftarrow \theta_0$ \;
	\While{not converged}{
		Sample batch of $( {s},  {a}, f( {s},  {a}))$ data \;
		Estimate $\nabla_\theta L(\theta, \lambda), \nabla_\lambda L(\theta, \lambda)$ using data \;
		$\theta^+ = \theta - \alpha_\theta \nabla_\theta L(\theta, \lambda)$ \;
		$\lambda \leftarrow \lambda + \alpha_\lambda \nabla_\lambda L(\theta, \lambda)$ \;
		$\lambda \leftarrow max(\lambda, 0)$ \;
		$\theta \leftarrow \theta^+$
	}
	\Return $\theta$ \;
	\caption{ModelLearning($\theta_0, D$)}
\end{algorithm2e}
Note that gradient ascent and descent steps can be modified to include momentum or include more complicated algorithms such as ADAM~\cite{kingma2014adam}.   In the next section we show that the sufficiently accurate learning formulated in \eqref{eq:constrained_model_learning} has almost zero duality gap which motivates the use of the primal-dual algorithm to solve it. One modification that many model learning methods use is to run the model while it is training to gather new data that is added to the training set. This is a DAGGER-like \cite{ross2011reduction} approach used by many methods \cite{Williams-ICRA-17, nagabandi2018neural}. 

\subsection{Almost Zero Duality Gap}\label{sec_zero_duality_gap}
By definition of the dual problem, it follows that the dual solution $D^*_\theta$ is always a lower bound for the primal solution $P^*_\theta$. That is, $P^*_\theta \geq D^*_\theta$ \cite{boyd2004convex}. The converse is however not true, but we can show that in the case of sufficiently accurate learning the duality gap is small. To that end we consider the following generalization of the problem 
\eqref{eq:constrained_model_learning}
\begin{align}
\label{eq:opt_problem_func}
P^* &= \min_{\phi \in \Phi} \mathbb{E}_B [l( {s},  {a}, \phi)] \\
& \text{s.t. } \mathbb{E}_B [g( {s},  {a}, \phi)] \leq 0, \nonumber
\end{align}
where instead of optimizing the weights of a function approximator, the optimization is done over the space of all possible integrable functions $\Phi$. We now have the following result shown in~\cite{ribeiro2012optimal}.
\begin{theorem}
	Given the optimization problem in \eqref{eq:opt_problem_func}, if (i) the distribution $B$, is non-atomic, (ii) the inequality constraints define a compact region within $\Phi$, and (iii) there exists a strictly feasible solution $(\phi, \lambda)$ (Slater's condition) then the duality gap is zero.
\end{theorem}
With this result, a natural question to ask is how the parameterization of functions $\phi \in \Phi$ affects the duality gap. The following theorem from~\cite{eisen2018learning} describes sufficient conditions under which a proxy of the duality gap is bounded.
\begin{theorem}
	\label{thm:parameterized_gap}
	For the optimization problems~\eqref{eq:opt_problem_dual} and~\eqref{eq:opt_problem_func}, if,
	\begin{itemize}
		\item there exists a strictly feasible solution $(\phi, \lambda)$ to the primal problem \eqref{eq:opt_problem_func};
		\item the parameterization $\theta$ of the function space $\Phi$ is a universal approximator within error $\delta$, i.e.
		$\mathbb{E}_{D} \left\| \phi - \phi_\theta \right\| \leq \delta$ for some $\theta$ for all $\phi \in \Phi$;
		\item the loss function, $l( {s},  {a}, \phi)$ is  expectation-wise Lipschitz continuous, i.e., there exists a $K$ such that $\mathbb{E} \left\| l( {s},  {a}, \phi_1) -  l( {s},  {a}, \phi_2)\right\|_\infty \leq K \mathbb{E} \left\| \phi_1 - \phi_2 \right\|_\infty$ for all $\phi_1, \phi_2 \in \Phi$;
	\end{itemize}
	then the optimal parameterized dual value $D_{\theta}^*$ is bounded by
	\begin{equation}\label{eq:dual_gap}
	P^* \leq D_\theta^* \leq P^* + \| {\lambda}^* \|_1 K \delta,
	\end{equation}
	where $\lambda^*$ is the solution of \eqref{eq:opt_problem_dual}.
\end{theorem}
The following proposition formalizes that the problem of sufficient accurate learning in \eqref{eq:constrained_model_learning} has small duality gap. 

\textbf{Proposition 1.} \textit{Sufficiently Accurate Learning and its dual, defined in \eqref{eq:constrained_model_learning_example},  satisfy the assumptions of Theorem 2. Hence \eqref{eq:dual_gap} holds for this case. }

\begin{proof}
First we look at the Lipschitz condition.
\begin{align*}
    & \mathbb{E} \left\| l(\boldsymbol{s}, \boldsymbol{a}, \phi_1) - l(\boldsymbol{s}, \boldsymbol{a}, \phi_2) \right\|_\infty \\
    & \leq \mathbb{E} \left\| \frac{\mathbb{I}_{\mathcal{K}}}{| f(\boldsymbol{s}, \boldsymbol{a})| }\right\| \| | \phi_1(\boldsymbol{s}, \boldsymbol{a}) - f(\boldsymbol{s}, \boldsymbol{a}) | - \\
    & \qquad \qquad \qquad \qquad | \phi_2(\boldsymbol{s}, \boldsymbol{a}) - f(\boldsymbol{s}, \boldsymbol{a}) | \|_\infty \\
    &\leq   \frac{1}{\min_{(\boldsymbol{s}, \boldsymbol{a})} | f(\boldsymbol{s}, \boldsymbol{a})| }\mathbb{E} \left\| \phi_1(\boldsymbol{s}, \boldsymbol{a}) - \phi_2(\boldsymbol{s}, \boldsymbol{a}) \right\|_\infty \\
    &= \frac{1}{\epsilon} \mathbb{E} \left\| \phi_1(\boldsymbol{s}, \boldsymbol{a}) - \phi_2(\boldsymbol{s}, \boldsymbol{a}) \right\|_\infty 
\end{align*}
Thus, the loss is expectation-wise Lipschitz continuous.
A strictly feasible solution, $\phi$, exists since the ground truth model, which is representable in $\Phi$, is strictly feasible.
Lastly, the parameterization used is the class of all neural networks which are universal function approximators. 
Therefore the conditions for Theorem \ref{thm:parameterized_gap} are fulfilled.
\end{proof}
This theorem states that for a large enough neural network, the gap between the optimal solution with no function approximation and the optimal solution to the parameterized dual problem scales with $\epsilon$. While this does not mean that the optimal parameterized dual problem can be solved, it does motivate the use of the primal dual method. A sample training curve is shown in Fig. \ref{fig:model_learning} for solving Eq. \ref{eq:constrained_model_learning_example}. More details about the specific training parameters are given in Section \ref{sec:experiments}.

\section{Control with Learned Models}

\begin{algorithm2e}[h]
	\label{alg:controller}
	\SetAlgoLined
	\KwIn{$\theta$, trained forward model parameters\\
		\quad \quad \quad $ {s}$, state to compute action for}
	$\lambda \leftarrow \lambda_0$ , set dual variable to an initial value \;
	$ {a} \leftarrow  {a}_0$, set action to some initial action \;
	\While{\text{not converged}}{
		${a}^+ =  {a} - \alpha_{ {a}} \nabla_{ {a}} L( {a}, \lambda)$\;
		$\lambda \leftarrow \lambda + \alpha_\lambda \nabla_\theta L( {a}, \lambda)$\;
		$\lambda \leftarrow \max(\lambda, 0)$ \;
		${a} \leftarrow {a}^+$ \;
	}
	\Return $ {a}$ \;
	\caption{Model Based Controller, $\pi( {s})$}
\end{algorithm2e}
The previous section describes a method for learning a sufficiently accurate model for controller design. In this section we describe how the learned model can be used to that end. For many planning algorithms such as A* or RRT \cite{lavalle2001randomized}, the model can simply be used to generate motion primitives that are more accurate. For optimization based planners, there is a variety of ways to use the model. One such method is to write out costs that explicitly include the model.

A deterministic policy is a function from the state to the action space $\pi:\mathcal{S}\to \mathcal{A}$.  A specific way to describe a desired policy is to minimize some cost, $c( {s},  {a}, \phi_\theta)$ associated with the performance of the system. This could be for instance the difference between the predicted state of the model and some desired state. The action selected has to typically satisfy some constraints imposed by the model, e.g., the action is bounded by the maximum torque of the motor in a robotic system, or obstacles in an environment. Denote by $g_c:\mathcal{S}\times\mathcal{A}\times\mathbb{R}^N\to \mathbb{R}^M$, the constraints imposed to the system and define the following policy 
\begin{align}
\pi( {s,\theta}) &= \argmin_{ {a}} c_f(s_T) + \sum_{t=1}^{T-1} c_s( {s_t},  {a_t}) \\
&\quad \text{s.t.  } g_s( {s_t},  {a_t}) \leq 0 \text{ for } t=1, \hdots, T-1 \nonumber \\
&\quad \quad g_f(s_T) \leq 0 \nonumber
\end{align}
where $c_f$ is the cost on the final state and $c_s$ is the cost on each step. In addition, $s_{t+1} = \phi_\theta(s_t, a_t)$. Thus $s_T$ will be the model $\phi_\theta$ applied $T - 1$ times to $s_1$.
Observe that the policy depends on the learned model, and thus it is also a function of the parameters $\theta$. In particular, if the residual error of learning the model dynamics is low, we can expect good performance out of such policies. 

Since, the model is a neural network, it is easy to obtain gradients $\nabla_{ {a}} \phi_{\theta}( {s},  {a})$ which can be used for the same primal-dual method mentioned in Sec. \ref{Methodology}. The only difference is that instead of optimizing model weights, the solver is optimizing for the inputs. This procedure is shown in Algorithm \ref{alg:controller} where the Lagrangian is defined as
\begin{align}
L( \boldsymbol{a}, \boldsymbol{\lambda})& = c_f(s_T) + \sum_{t=1}^{T-1}c_s( {s_t},  {a_t}) + \\
& \lambda_0 g_f( {s_T}) + \sum_{t=1}^{T-1} \lambda_t g_s(s_t, a_t) \nonumber
\end{align}
Learning a model and using such an optimization problem as the policy allows different controllers to be designed for different goals and constraints. This optimization problem can be solved repeatedly, taking only one action each time in a Model Predictive Control framework. Another alternatives can be to formulate optimization problems where the dynamics are constraints as in direct collocation \cite{kelly2017introduction}.

\section{Experiments}\label{sec:experiments}
We apply the sufficiently accurate model learning framework to a robot arm bouncing a ball with a paddle, to demonstrate its effects. The paddle has 5 degrees of freedom: the position in the three dimensional space and the pitch and roll angles. We model the impact of the ball on the paddle. It takes as input the velocities of the paddle and the ball before they collide as well as the orientation of the paddle and outputs the velocity of the ball after collision. 
The controller is then tasked to solve the optimization problem
\begin{align}\label{eq:ball_paddle_opt}
\pi( {s}) =& \min_{ {a}} \left\| p(\phi_\theta( {s,a})) -  {p}_{desired} \right\| \\
& \text{s.t. }  {roll}_{min} \leq  roll \leq  {roll}_{max} \nonumber \\
& \quad {pitch}_{min} \leq  pitch \leq  {pitch}_{max} \nonumber \\
& \quad v_{min} \leq \left\| v_{rel} \right\| \leq v_{max} \nonumber \\
& \quad \quad p(\phi_\theta(s,a)) \not \in P_{obstacle} \nonumber,
\end{align}
where $v_{rel}$ is the relative velocity of the ball to the paddle, and $p(\phi_\theta( {s,a}))$ is a function that outputs where on the xy plane the ball will hit next. The actions, $a$, consist of a chosen roll and pitch angle as well as desired paddle velocity. The goal of the controller is to bounce a ball at some specific xy location, $p_{desired}$ while obeying action constraints and not hit any obstacles. The simulation was created using libraries from the DeepMind Control Suite \cite{tassa2018deepmind}.

While simulation has drawbacks such as its inability to accurately capture full noise characteristics of a system, it is a useful tool to test out model learning. We can inject known errors into a model and compare how our method does with the real model. This is difficult or impossible to do in real life. 

We will first present experiments in which we learn a full model of the system. That is, the neural network is tasked to output the full model, $f(s, a)$. Next, we will present experiments where a residual model is learned. In the residual model, the neural network is tasked to output the residual $f(s, a) - \hat{f}(s, a)$ where $\hat{f}$ is a (possibly wrong) analytic model. This is the situation where we may have an initial guess of what the model of the system looks like from physical measurements but can be fine-tuned with data.

\begin{figure}[h]
	\centering
	\includegraphics[width=0.475\textwidth]{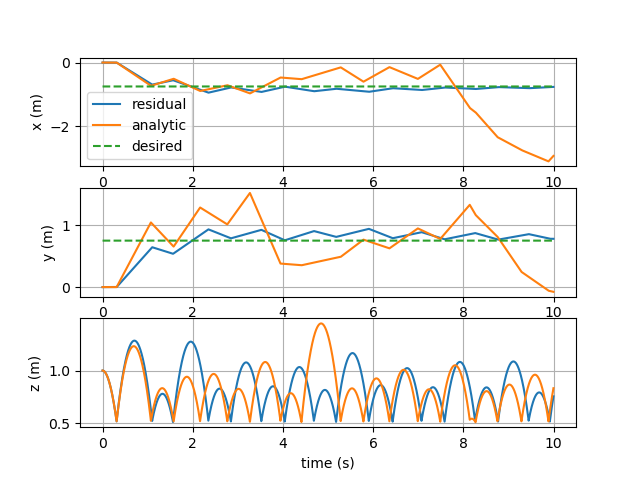}
	\caption{\textbf{Analytic vs Residual Model.} The (x, y z) trajectory of the ball is plotted for both the a wrong analytic model as well as the residual model learned on top of it. The analytic model has an error of 0.2 radians in its observation of the roll angle of the paddle. This leads to poor performance by the optimizing controller. The residual model corrects these errors and can track the desired (x, y) location quite well.}
	\label{fig:trajectory}
\end{figure}

\begin{figure}[h]
	\centering
	\includegraphics[width=0.475\textwidth]{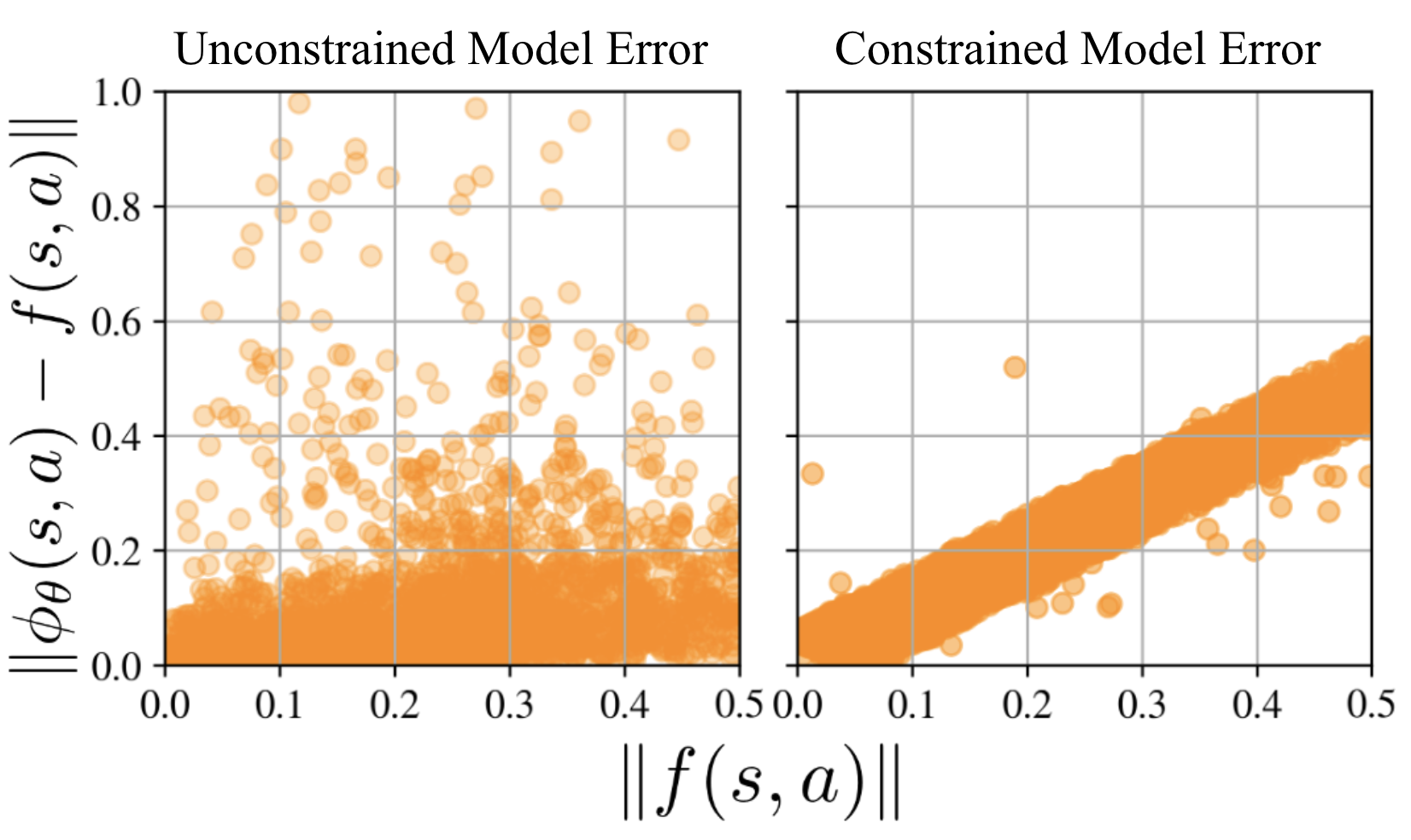}
	\caption{\textbf{Model errors vs. state magnitude}. On the left side, data from a model trained with Eq. \ref{eq:model_learning} is shown. The right side shows data from a model using Eq. \ref{eq:constrained_model_learning_example}. The scatter plot shows the errors of running each model on a validation set. The unconstrained problem leads to a good loss in expectation, however the errors are distributed poorly. There might be large errors for states with a small magnitude. The normalized loss on the right allows the large states to have large error and small states to have small error.}
	\label{fig:model_errs}
\end{figure}
\subsection{Full Model Learning}
To train the full model, we first collect a dataset from running the controller described by Eq. \ref{eq:ball_paddle_opt} where we replace $\phi_\theta$ with a faulty analytic model on the system. The faulty analytic model is mistaken about the roll angle at which the robot is holding the paddle (off by 0.1 radians). While this level of angular error can be hard to measure, it has a large effect on how the ball bounces. We collect data while simulating the system for the equivalent of 42 real world minutes. The analytic model used for the velocity of the ball after impact is given as
\begin{equation}
    \boldsymbol{v}_{ball} = \alpha * (\boldsymbol{v}_{rel} - 2 \boldsymbol{n} (\boldsymbol{n} \cdot \boldsymbol{v}_{rel})) + \boldsymbol{v}_{paddle}
\end{equation}
where $\boldsymbol{v}_{rel}$ is the relative velocity of the ball to the paddle, $\alpha$ is a coefficient of restitution, and $\boldsymbol{n}$ is the normal vector representing the orientation of the paddle. The normal vector is observed with an error.

First, we examine the distribution of errors, comparing with a simple model learning approach where just the loss $\mathbb{E}[ l(s, a, \phi_\theta)]$ is minimized with the objective defined in Eq. \ref{eq:constrained_model_learning_example}. They will be denoted as unconstrained and constrained models, respectively. The neural network model that is used is a simple 2 hidden layer fully connected neural net with 128 neurons in each hidden layer with parametric rectified linear activations \cite{he2015delving}. The network was trained using the ADAM optimizer with an initial learning rate of 1e-3. For the constrained example, both $\delta$ and $\epsilon$ were chosen to be 0.1. The results in Fig. \ref{fig:model_errs} show that the model trained with the sufficiently accurate objective has a different error distribution.

One hypothesis that can be drawn from this figure is that the gradients of the model may be less noisy as there are not sudden jumps in error. For optimization based controllers and planners, this is a big benefit. To see what effect this has on the controller, the controller is run with each model 500 times with different goal locations, $p_{desired}$ as well as different velocity constraints, $v_{min}$ and $v_{max}$. $p_{desired}$ is uniformly distributed  in the region $\{(x, y) | -1m \leq x \leq 1m, -1m \leq y \leq 1m\}$, $v_{min}$ uniformly sampled from the interval $[3m/s, 4m/s)$, and $v_{max}$ is selected to be above $v_{min}$ by between $1m/s$ to $2m/s$. Of these 500 experiments, we consider it a failure if the ball falls off the paddle, or when it is hit far away and can not recover. The ``Full Model" rows of Table \ref{table:model_errs} shows that the failure rate of the sufficiently accurate model is much lower as well as having a lower mean error when it does not fail. The mean error is the mean position error (to the desired location) over time. It is a crude measure of both how fast the controller gets to the goal and how well it stays on it.

\begin{table}[h]
	\centering
	\caption{\textbf{Controller performance with learned models}}
	\begin{tabular}{c|c|c|c|}
		\cline{2-4}
		& & Unconstrained & Constrained \\
		\cline{2-4}
		\multirow{2}{*}{Full Model}&Failure &$20.8 \%$ & $0.8 \%$\\ \cline{2-4}
								   &Mean error & $0.3136$ & \textbf{0.2124} \\
								   \cline{2-4}
								   \cline{2-4}
		\multirow{2}{*}{Residual Model}& \cellcolor{black!20} Failure & \cellcolor{black!20}$0 \%$ & \cellcolor{black!20} $0 \%$\\  \cline{2-4}
								   & \cellcolor{black!20} Mean error & \cellcolor{black!20} $0.164$ & \cellcolor{black!20} \textbf{0.156} \\
								   \cline{2-4}
	\end{tabular}

	\label{table:model_errs}
\end{table}

\balance
\subsection{Residual Model Learning}
In many scenarios, we will have a base analytic model that may be wrong, but would like to improve upon it rather than learning a full model from scratch. Using the same dataset as described in the previous section, we can train a residual model using the unconstrained form in Eq. \ref{eq:model_learning} and a constrained model. All architecture details and hyperparameters used in the residual model are the same as in the full model training.  With the residual models, there are no failures as the base analytic model performs well enough to prevent that. 
The mean error rates are shown in Table \ref{table:model_errs}. Trajectories of the ball using the analytic and the residual constrained model is shown in Fig. \ref{fig:trajectory}. We can see with a wrong analytic model, the controller does not track the desired position well. As expected, it has a consistent bias to overshoot when trying to correct its position and ends up with a jagged trajectory. The constrained residual model tracks the position much better and does not deviate as the analytic model does.

\begin{figure}[h]
	\centering
	\includegraphics[width=0.475\textwidth]{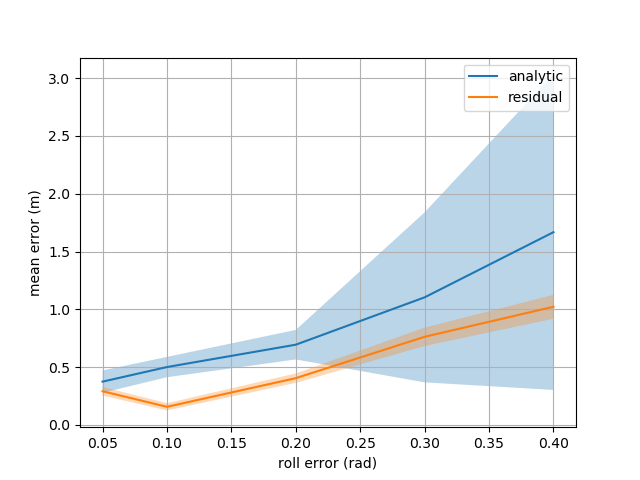}
	\caption{\textbf{Analytic vs Residual Model.} A residual model is trained using the same dataset for the same number of iterations with different base analytic models of increasing error. Both the analytic and constrained residual models are then evaluated by running them for 50 different task goals and constraints and their mean errors are computed. The transparent regions show half of a standard deviation above and below the mean.}
	\label{fig:increasing_roll}
\end{figure}

We compare the constrained residual model with different analytic models with varying levels of error in Fig. \ref{fig:increasing_roll}. The error is computed the same way as in Table \ref{table:model_errs}. The residual model has both a lower mean error as well as having a much tighter variance, which means it has more consistent performance. 

\section{Discussion and Conclusion}\label{sec:discussion}
In looking at model learning, we have seen that using the Sufficiently Accurate formulation can bring better results simply by changing the optimization objective. We believe this is due to the fact that this formulation can smooth out error characteristics and provide better gradients for the controller to work with. This methodology is orthogonal with most other model learning algorithms as it makes a suggestion to use constraints as a way to control the errors and gradients.

There are several drawbacks of this method. One of which is that computing derivatives of the model through a neural network can be computationally expensive. 
This makes it more difficult to deploy on systems that require fast control loops.
This can possibly be alleviated by training a fast policy for specific tasks by imitating the more expensive model based solution. 
Another drawback is that $\epsilon_i$ in Eq. \ref{eq:constrained_model_learning} is determined by hand. These may need to be adjusted if the problem is infeasible or not tight enough.

There are several branches of future exploration. One is to implement this on a robot arm to test how this method can handle other types of errors that can occur. Another is to explore different types of constraints on different portions of the state space, or an automated way to choose constraints. Testing how different controllers and planners interact with these models can inform us of other characteristics of models that may be important to study.

\section*{Acknowledgments}
The authors would like to thank Pratik Chaudari for valuable conversations as well as funding by NSF Grant No. DGE-1321851 and the Intel Science and Technology Center for Wireless
Autonomous Systems. 
Any opinions, findings, and conclusions do not necessarily reflect the views of the NSF.

\newpage
\bibliographystyle{ieeetr}
\bibliography{bib}
\end{document}